\def\paperID{}
\def\confName{CVPR}
\def\confYear{2026}

\def\paperTitle{
EasyOmnimatte: Taming Pretrained Inpainting Diffusion Models for \\ End-to-End Video Layered Decomposition 
}

\newcommand{\name}{EasyOmnimatte\xspace}

\def\authorBlock{
    Yihan Hu$^{1}$ \qquad
    Xuelin Chen$^2$ \qquad
    Xiaodong Cun$^{1, \dagger }$ \\ \\
    $^1$~GVC Lab, Great Bay University \qquad
    $^2$~Adobe Research \\
     \\ \url{https://github.com/GVCLab/EasyOmnimatte}
}

\newif\ifreview 
\newif\ifarxiv \newcommand{\arxiv}{\arxivtrue}
\newif\ifcamera 
\newif\ifrebuttal

\makeatletter
\DeclareRobustCommand\onedot{\futurelet\@let@token\@onedot}
\def\@onedot{\ifx\@let@token.\else.\null\fi\xspace}

\def\eg{\emph{e.g}\onedot} 
\def\ie{\emph{i.e}\onedot}

\makeatother

\arxiv 

\pdfoutput=1
\documentclass[10pt,twocolumn,letterpaper]{article}
\usepackage{xcolor}

\definecolor{cvprblue}{rgb}{0.21,0.49,0.74}
\definecolor{lightyellow}{RGB}{252, 252, 233}
\definecolor{beaublue}{RGB}{240, 247, 255}

\ifreview \usepackage[review]{cvpr} \fi
\ifarxiv \usepackage[pagenumbers]{cvpr} \fi
\ifrebuttal \usepackage[rebuttal]{cvpr} \fi
\ifcamera \usepackage{cvpr} \fi

\usepackage{graphicx}
\usepackage{amsmath}
\usepackage{amssymb}
\usepackage{booktabs}
\usepackage{xspace}
\usepackage{bm}

\usepackage{times}
\usepackage{microtype}
\usepackage{epsfig}
\usepackage[table,xcdraw]{xcolor}
\usepackage{caption}
\usepackage{float}
\usepackage{placeins}
\usepackage{color, colortbl}
\usepackage{stfloats}
\usepackage{enumitem}
\usepackage{tabularx}
\usepackage{xstring}
\usepackage{multirow}
\usepackage{xspace}
\usepackage{url}
\usepackage{subcaption}
\usepackage{xcolor}
\usepackage[hang,flushmargin]{footmisc}
\ifcamera \usepackage[accsupp]{axessibility} \fi

% ...

\ifarxiv  \fi

\newcommand{\R}[1]{{%
    \textbf{%
        \ifstrequal{#1}{1}{\textcolor{red}{R#1}}{%
        \ifstrequal{#1}{2}{\textcolor{blue}{R#1}}{%
        \ifstrequal{#1}{3}{\textcolor{magenta}{R#1}}{%
        \ifstrequal{#1}{4}{\textcolor{teal}{R#1}}{%
                           \textcolor{cyan}{R#1}%
        }}}}%
    }%
}}

\usepackage{xr-hyper}

\makeatletter
\newcommand*{\addFileDependency}[1]{
  \typeout{(#1)}
  \@addtofilelist{#1}
  \IfFileExists{#1}{}{\typeout{No file #1.}}
}

\makeatother

\usepackage[pagebackref,breaklinks,colorlinks]{hyperref}
\usepackage[capitalize]{cleveref}
\crefname{section}{Sec.}{Secs.}
\crefname{table}{Table}{Tables}
\crefname{figure}{Fig.}{Figs.}

\definecolor{alizarin}{rgb}{0.82, 0.1, 0.26}

\hypersetup{
    colorlinks = true,
    linkbordercolor = {white},
    citecolor=cvprblue,
    urlcolor=alizarin
}

\begin{document}
\title{\paperTitle}
\author{\authorBlock}

\twocolumn[{
\maketitle
\begin{center}
    \captionsetup{type=figure}
    \vspace{-2em}
\includegraphics[width=1.\textwidth]{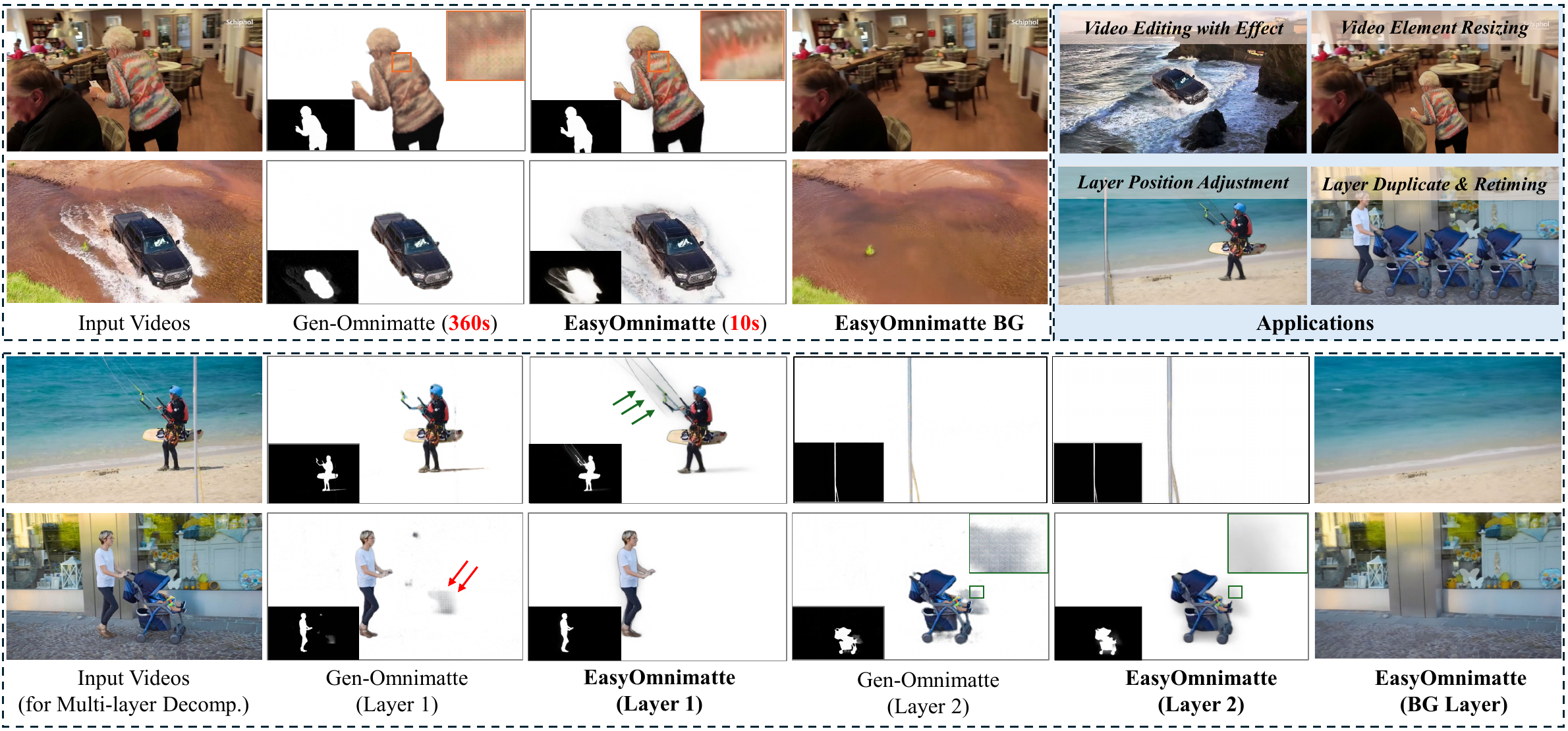}
    \vspace{-2em}
    \captionof{figure}{
    \textit{\textbf{\name}} is \emph{the first, end-to-end video omnimatte} method, eliminating the need for slow, multi-stage optimization-based pipelines (e.g., Gen-Omnimatte~\cite{lee2025generative}).
    \name produces high-fidelity alpha mattes that accurately capture the associated effects, \emph{all within just seconds}, 
    representing a significant gain over existing state-of-the-art methods in both quality and efficiency.
    }
    \label{fig:teaser}
\end{center}
}]

\begin{abstract}
Existing video omnimatte methods typically rely on slow, multi-stage, or inference-time optimization pipelines that fail to fully exploit powerful generative priors, producing suboptimal decompositions.
Our key insight is that, if a video inpainting model can be finetuned to remove the foreground-associated effects~\cite{lee2025generative}, then \text{it must be inherently capable of perceiving these effects}, and hence can also be finetuned for the complementary task: foreground layer decomposition with associated effects.
However, although naïvely finetuning the inpainting model with LoRA applied to all blocks can produce high-quality alpha mattes, it fails to capture associated effects.
Our systematic analysis reveals this arises because effect-related cues are primarily encoded in specific DiT blocks and become suppressed when LoRA is applied across all blocks.
To address this, 
we introduce \name, 
the first unified, end-to-end video omnimatte method.
%
%
%
Concretely, we finetune a pretrained video inpainting diffusion model to learn dual complementary experts while keeping its original weights intact: an Effect Expert, where LoRA is applied only to effect-sensitive DiT blocks to capture the coarse structure of the foreground and associated effects, and a fully LoRA-finetuned Quality Expert learns to refine the alpha matte.
During sampling, Effect Expert is used for denoising at early, high-noise steps, while Quality Expert takes over at later, low-noise steps. This design eliminates the need for two full diffusion passes, significantly reducing computational cost without compromising output quality.
Ablation studies validate the effectiveness of this Dual-Expert strategy.
Experiments demonstrate that \name sets a new state-of-the-art for video omnimatte and enables various downstream tasks, significantly outperforming baselines in both quality and efficiency. 
\end{abstract}
\section{Introduction}
\label{sec:intro}

Decomposing videos into layered representations is a fundamental problem in computer vision,
with significant applications in video editing and visual effects. 
Effective decomposition enables a variety of video editing tasks, including object removal, background replacement, and other creative workflows.
However, extracting accurate alpha mattes for target foregrounds is labor-intensive, especially for complex effects arising from object-environment interactions such as shadows, reflections, generated smoke, or splashes.
Apparently, this task is highly difficult to achieve using conventional methods.
To address this, the pioneering work~\cite{lu2021omnimatte} has formulated the \emph{Omnimatte} task, which seeks to separate a foreground object and all its associated visual effects into a single coherent layer, and trains a UNet model on the input video in a self-supervised manner.
Subsequently, there has been a surge of follow-up omnimatte methods improving it from various perspectives.
%
However, these follow-ups typically rely on slow, multi-stage, or inference-time optimization pipelines that fail to fully exploit powerful generative priors, producing suboptimal decompositions.

In this work, we present the \emph{first unified}, \emph{end-to-end} framework for video omnimatte generation, addressing the aforementioned limitations.
%
Our key insight is that, if a video inpainting model can be finetuned to remove the foreground-associated visual effects presented with the original, unmasked video~\cite{winter2024objectdrop,yu2025objectmover,yu2025omnipaint, lee2025generative}, 
which is a task complementary to video omnimatte,
then \emph{the model must inherently possess the ability to perceive these effects}. Consequently, the same model can be finetuned to include such effects in the foreground layer prediction.
To this end, we explore employing lightweight Low-Rank Adaptation (LoRA)~\cite{hu2022lora} modules to finetune the inpainting model on a synthetic matting dataset, enabling it to directly predict the foreground layer in a single, efficient stage.
Nonetheless, our initial experiments with this straightforward setup revealed a critical and unexpected challenge: while the finetuned model effectively produced high-quality alpha mattes for the foreground objects, it failed to consistently perceive and capture associated effects such as shadows and reflections — even when trained with ground-truth targets that include them.

Intrigued by this phenomenon, we conduct a systematic analysis of the model's internal mechanisms. 
Inspired by ObjectDrop~\cite{winter2024objectdrop} and the analysis in Gen-Omnimatte~\cite{lee2025generative}, we examine how different blocks of the inpainting model respond to subtle “effect” cues.
Our structural analysis of the inpainting model reveals that the perception of these effects is concentrated within a specific subset of blocks in the DiT architecture. 
Applying LoRA to all DiT blocks tends to overemphasize the main alpha prediction task and inadvertently suppresses the weaker, effect-related signals, manifested as missing effects in the omnimatting results.
With these findings,
our key idea is to finetune a pretrained video inpainting diffusion model to learn two complementary experts while keeping its original weights frozen: an \emph{Effect Expert}, where LoRA is applied only to effect-sensitive intermediate stages to capture the \emph{coarse}, coarse structure of the foreground and its associated effects, and a \emph{Quality Expert}, where LoRA is applied to all DiT blocks, learns to add \emph{fine} details to the alpha matte.
%
Notably, during diffusion sampling, the Effect Expert model is only employed at early, high-noise stages to generate coarse, effect-aware omnimatte predictions, while the Quality Expert model refines the alpha matte only at later, low-noise stages.
Rather than running the two finetuned models sequentially for full diffusion sampling, this alternating strategy achieves high-quality results with roughly half the computational cost.
%
Experiments confirm that this classic \emph{coarse-to-fine} strategy, which leverages both the specialized and full LoRA adaptations, is crucial for producing alpha mattes that are accurate and faithfully preserve associated effects. Our evaluation demonstrates that \name sets a new state-of-the-art (SOTA) video omnimatte. Importantly, the proposed method maintains an end-to-end, feed-forward decomposition of the foreground layers, ensuring both effectiveness and efficiency.

In summary, our main contributions are as follows:
\begin{itemize}
\item We introduce the \emph{first end-to-end video omnimatte} method, which directly adapts a pre-trained video inpainting model to capture associated visual effects, eliminating the need for slow, multi-stage optimization-based pipelines.
\item We perform a structural analysis revealing that the standard inpainting pretraining objective inherently conflicts with the preservation of foreground effects, strongly motivating our key designs.
\item 
We finetune a pretrained video inpainting diffusion model to learn two specialized experts. These complementary experts operate in tandem across the high- and low-noise denoising stages,
producing high-fidelity alpha mattes that accurately capture the associated effects, all without additional computational overhead.
\end{itemize}

\section{Related Work}
\label{sec:related}

\noindent\textbf{Foreground Isolation and Matting.} The most straightforward way to extract the foreground layer from a video is via matting~\cite{porter1984compositing, wang2008image}. Converting the foreground into a transparent (RGBA) format enables a wide range of video editing tasks, making video matting a fundamental technique for such applications.
In recent years, deep learning-based matting methods for both images~\cite{xu2017deep,yu2021mask,yao2024vitmatte,hu2024diffusion,wang2024matting} and videos~\cite{sun2021deep,lin2022robust,huynh2024maggie,yang2025matanyone} have achieved significant improvements, enhancing the accuracy of alpha matte prediction while reducing the need for extensive guidance information.
These methods, often trained on fine-grained labeled datasets, can automatically learn image features to predict high-quality alpha mattes. However, limitations in the diversity of training data \cite{xu2017deep,qiao2020attention,sun2021deep,lin2021real} and specific training strategies \cite{ke2022modnet,lin2022robust,huynh2024maggie} often limit their performance to certain foreground categories, most notably humans. 
Furthermore, they tend to fail to capture foreground-associated effects, such as shadows or reflections. 
While some methods \cite{sengupta2020background,lin2021real} leverage a known background to partially address these limitations, they impose strict requirements, such as a static scene or a pre-captured clean background plate.

\noindent\textbf{Video Matting with Associated Effects.} 
A series of methods, exemplified by Omnimatte \cite{lu2020layered,lu2021omnimatte}, enables the decomposition of a video into foreground and background layers, with the foreground layer encompassing associated visual effects. 
These approaches aim to completely separate the foreground layer from the video, leveraging motion cues through flow-based techniques \cite{adelson1995layered,brostow1999motion,wang1994representing} to achieve this effect. Subsequent works have further enhanced the capabilities of Omnimatte by incorporating deep image priors \cite{lu2022associating,gu2023factormatte}, extending planar homography constraints through non-rigid warping \cite{lu2020layered,ye2022deformable} or 3D scene representations \cite{suhail2023omnimatte3d,lin2023omnimatterf}. However, a significant drawback is the reliance on restrictive motion assumptions, which can result in severely degraded performance when such assumptions no longer hold. 

\noindent\textbf{Video Inpainting for Object Removal.} 
Both background matting and Omnimatte-like optimization methods rely on precise background guidance. 
On the other end, the rapid progress of visual generative models~\cite{ho2022imagen,blattmann2023stable,girdhar2023emu,liu2024sora,chen2024videocrafter2,yang2024cogvideox,hacohen2024ltx,kong2024hunyuanvideo,wan2025wan} has greatly advanced video inpainting techniques, enabling accurate background layer prediction.
Early video inpainting methods focused on natural content completion by exploiting spatio-temporal cues from adjacent frames~\cite{wang2019video,chang2019free,liu2021fuseformer,zhou2023propainter} or propagating optical flow~\cite{li2022towards,zhang2022flow} to reduce hallucinations and artifacts within masked regions.
More recent approaches \cite{li2025diffueraser,zi2025minimax} directly adapt and fine-tune large video generative models for efficient object removal. 
Nonetheless, these methods cannot eliminate the associated effects of an object, often leaving residual artifacts inside the inpainted background. 
In image object removal, several recent methods \cite{winter2024objectdrop,yu2025objectmover,yu2025omnipaint} have specifically addressed the challenge of removing both objects and their associated effects. 
The key idea behind these approaches is to condition the generative model on the original, unmasked image, enabling it to learn the correlations between objects and their corresponding effects.
Built upon this success, Generative Omnimatte~\cite{lee2025generative} extended it to video generation models, successfully training a video removal model capable of cleanly removing objects along with their associated visual effects.

While recent methods can estimate the background by removing objects, using this pseudo-background for foreground separation still remains challenging. 
On one hand, per-frame techniques, such as background matting, often exhibit temporal inconsistencies like flickering and struggle to capture foreground-associated effects. On the other hand, two-stage approaches like Generative Omnimatte are not only computationally expensive due to test-time optimization but also prone to error propagation, as inaccuracies in background generation can corrupt the final foreground decomposition.
In this work, we propose the first unified framework that simultaneously predicts the background layer and separates the foreground along with its associated effects, all within a single end-to-end model.
\section{Method}
\label{sec:method}

\newcommand{\diff}{\ensuremath{\mathcal{G}}\xspace}
\newcommand{\fg}{\ensuremath{\bm{F}}\xspace}
\newcommand{\bg}{\ensuremath{\bm{B}}\xspace}
\newcommand{\am}{\ensuremath{\bm{\alpha}}\xspace}


We tackle the vedio omnimatte task, which seeks to decompose an input video $\bm{V} \in \mathbb{R}^{N \times H \times W \times 3}$ into a foreground layer $\fg \in \mathbb{R}^{N \times H \times W \times 3}$, its corresponding alpha matte $\am \in [0, 1]^{N \times H \times W}$, and a restored background layer $\bg \in \mathbb{R}^{N \times H \times W \times 3}$. 
These layers must adhere to the standard alpha compositing equation: $\bm{V} = \am \odot \fg + (1 - \am) \odot \bg$, where $\odot$ denotes element-wise multiplication. In contrast to multi-stage pipelines~\cite{lu2021omnimatte,lee2025generative} that predict $\bg$ and $\{\am, \fg\}$ sequentially, we formulate this as a direct, end-to-end prediction problem with a single model $\diff$. This model, with trainable parameters $\Theta$, is trained to map the input video directly and simultaneously to the complete set of decomposition layers. This unified transformation is expressed as:
\begin{equation}
\label{eq:e2e_omnimatte}
(\fg, \am, \bg) = \diff(\bm{V},\bm{M}, \bm{c}; \Theta),
\end{equation}
where $\bm{M}$ indicates the per-frame coarse foreground masks (\ie, without the effects), and $\bm{c}$ the inpainting condition. 

\begin{figure*}[tp]
    \centering
    \includegraphics[width=\linewidth]{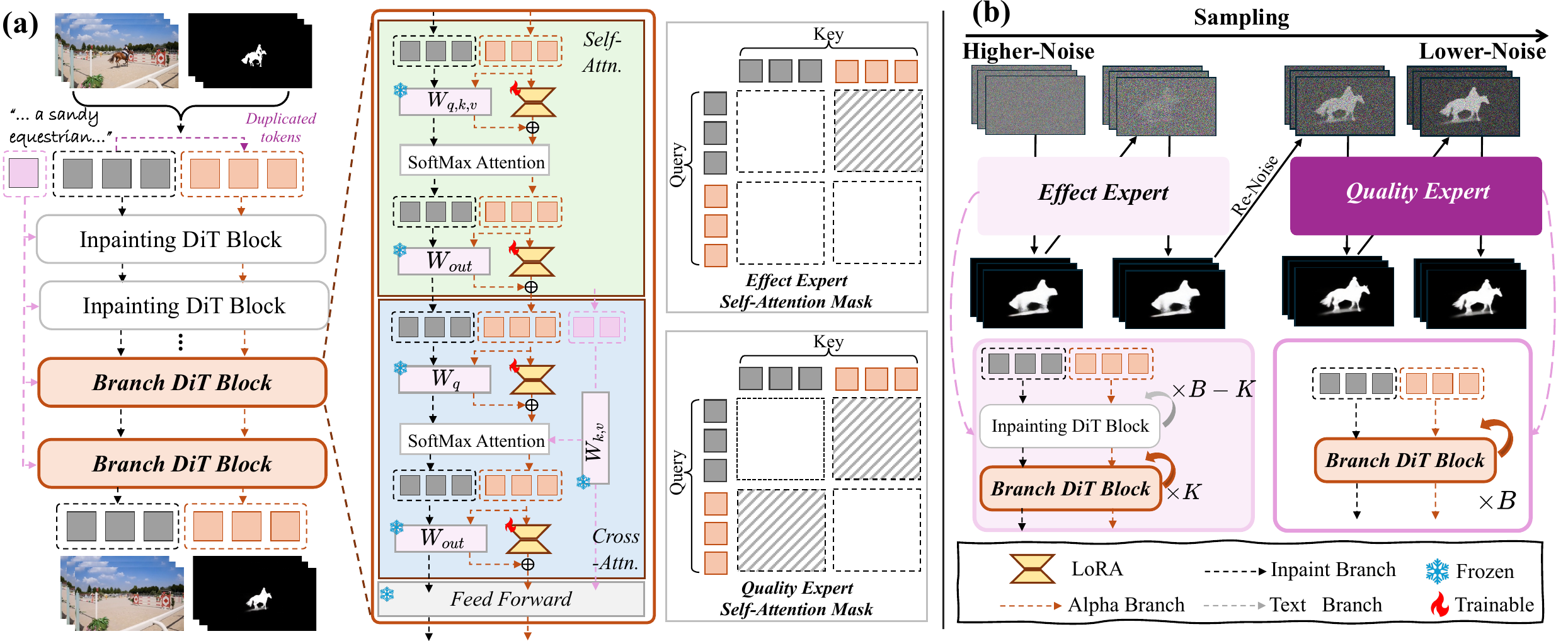}
    \vspace{-2em}
    \caption{\textit{\textbf{\name.}} 
    a) We \emph{branch out} LoRA-finetuned blocks from the original inpainting DiT blocks to jointly predict the alpha matte, alongside the pretrained model.
    In each Branch DiT Block, LoRA are applied only to the duplicated set of input tokens, leaving the original inpainting branch unaffected.
    b) During sampling, the \textit{Effect Expert} model is only employed at early, high-noise stages to generate coarse, effect-aware omnimatte predictions, while the \textit{Quality Expert} model refines the alpha matte only at later, low-noise stages.
    This alternating strategy achieves high-quality results with greatly reduced compute cost, compared to individually sampling.
    }
    \vspace{-1em}
    \label{fig:pipeline}
\end{figure*}

As shown in Fig.~\ref{fig:pipeline}, to mitigate the data sparsity issue, our core idea is to repurpose the powerful generative prior of a pre-trained video inpainting model in Sec.~\ref{sec:baseline}.
However, we observe that a naive fine-tuning of such a model struggles to capture subtle but crucial foreground effects (\eg, shadows, reflections). 
To understand this limitation, we first conduct a rigorous structural analysis (Sec.~\ref{sec:analysis}) of the inpainting model, revealing an innate functional conflict in its later blocks that actively suppresses effect-related signals. 
Guided by this key observation, we propose a novel Dual Experts Sampling ~(Sec.~\ref{sec:pipeline}) that resolves this conflict. 

\subsection{Repurpose Inpainting Diffusion for Omnimatte}
\label{sec:baseline}

Video inpainting models inherently encode a strong separation prior, allowing them to \emph{remove} foregrounds and their associated effects~\cite{lee2025generative}. We leverage this prior differently: by adapting the model to directly predict the foreground's alpha matte, we achieve a unified, end-to-end Omnimatte pipeline.
As Figure~\ref{fig:pipeline} shows, to achieve omnimatte prediction in a feed-forward way without compromising the model's powerful priors, we introduce \emph{Branch DiT Blocks}, constructed by adding trainable LoRA modules~\cite{hu2022lora} to the original DiT blocks, to recover the alpha mate with additional tokens.

The baseline inpainting model is conditioned on the complete video frames $\bm{F}$, per-frame foreground masks $\bm{M}$ provided by modern segmentors~\cite{ravi2024sam2}, and a brief environmental text description $c$ for background guidance. $\bm{F}$ and $\bm{M}$ are incorporated into the noise latents of the inpainting DiT through channel concatenation and projection to obtain the input visual token. Our method directly inherits this complete input scheme with additional tokens. Specifically, we first duplicate the input visual tokens as well as the rotary positional embeddings~(RoPE) following \cite{wang2025transpixeler}, then token-wisely concatenate them, creating two parallel sets: the original tokens destined for the inpainting task and a copied set for alpha matte prediction. These lightweight LoRA are applied \textit{exclusively} to the copied set of tokens. Consequently, the original, frozen backbone of the inpainting model operates on the untainted original tokens, proceeding with its background prediction task, while the LoRA-equipped branch processes the copied tokens, effectively redirecting their output from background prediction toward alpha matte estimation.

We choose to predict the alpha matte $\hat{\am}$ instead of the foreground layer ${\hat{\fg}}$~\cite{zhang2024transparent} since the alpha matte is well-defined across the entire frame~(\textit{i.e.}, it has a value of 0 in pure background regions), providing a more stable and well-posed prediction target. In contrast, the foreground pixels are only defined where the alpha is greater than zero.

Since we froze the original video inpainting model, the original input of the video inpainting model predicts the background $\hat{\bg}$, whereas the alpha matte $\hat{\am}$ is generated by the additional tokens, as shown in Fig.~\ref{fig:pipeline}. 
We then analytically recover the foreground layer $\hat{\fg}$ by rearranging the standard compositing equation. For each frame $f$, the foreground $\fg_f$ is computed as:
\begin{equation}
    \hat{\fg}_f = \frac{\text{clip}(\bm{I}_f - (1 - \hat{\am}_f) \cdot \hat{\bg}_f)}{\hat{\am}_f + \bm{\epsilon}},
    \label{eq:foreground_recovery}
\end{equation}
where $\bm{I}_f$ is the original frame, $\hat{\bg}_f$ is the inpainted background, and $\bm{\epsilon}$ is a small constant to prevent division by zero. The clip function ensures that the resulting pixel values remain within the valid [0, 1] range. 

\subsection{Block-Wise Analysis of Effects Association}
\label{sec:analysis}


Following the above structure, we observe that even when employing all \textit{Branch DiT} blocks, video inpainting models still fail to effectively capture and preserve the associated effects. Given the efficacy of inpainting models in erasing an object as well as its attendant effects that are not specified, we infer an underlying sequential process of first perceiving and then eliminating these phenomena. To validate this hypothesis, we perform a block-wise analysis to probe the mechanisms for effect-perception and processing at different depths of the network.

%


We calculate the \textit{Effects Association Score} \cite{winter2024objectdrop, lee2025generative} $s(p)$ for each foreground pixel $p$, based on a block's self-attention map $\mathbf{W}$. This score represents the fraction of a pixel's attention that is directed towards the effects region $\bm{M}^{e}$:
\begin{equation}
    s(p) = \frac{\sum_{y \in \bm{M}^{e}} \mathbf{W}_{p,y}}{\sum_{x \in I} \mathbf{W}_{p,x}},
    \label{eq:assoc_score}
\end{equation}
so for frame $f$, we can obtain the effect-related attentive map of the $b$-th DiT blocks, denoted as $\bm{S}_{f, b}$.

%
To quantify how different attention blocks in a DiT perceive associated effects, we compute a normalized, block-wise contribution score. 
Specifically, for each frame $f$, we extract the ground-truth effect mask $\mathbf{M}^{e}_f$ by taking the intersection of a binarized alpha matte and the complement of a dilated foreground mask, isolating the effect region from the object itself. Supplementary presents the full derivation.

\begin{figure}[tp]
    \centering
    \includegraphics[width=\linewidth]{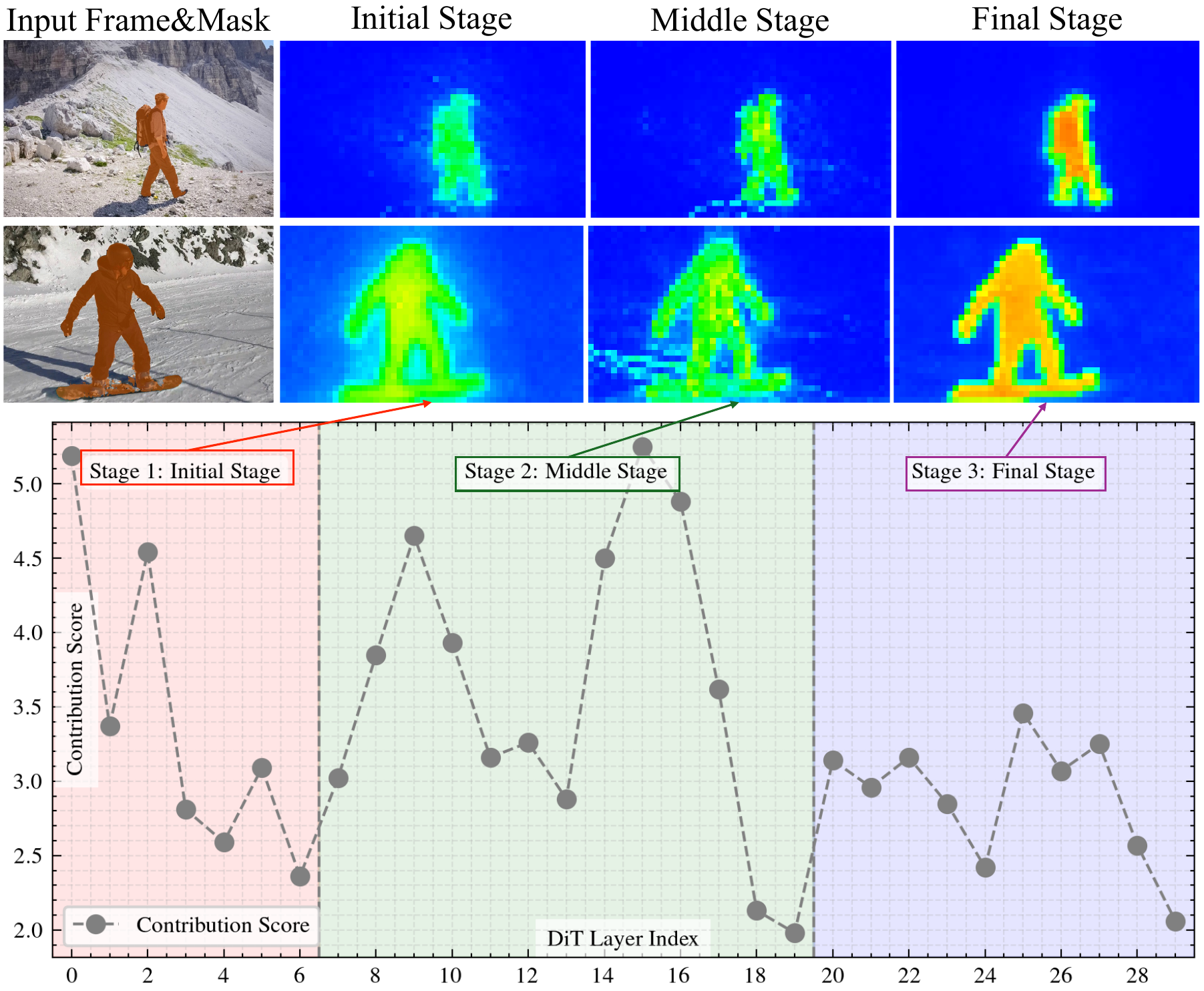}
    \vspace{-2em}
    \caption{
    \textbf{\textit{Our block-wise Analysis.}} 
    \textbf{Bottom}: Contribution scores per layer reveal the three inpainting stages. 
    \textbf{Top}: Averaged attention maps ($\bm{S}_b$) for each stage show that the middle stage is most sensitive to effects (e.g., shadows), which are then suppressed in the final stage, corroborating our perception-then-elimination hypothesis.
  }
    \vspace{-1.3em}
    \label{fig:analysis_plot}
\end{figure}

The block-wise contribution score ${C}_{b}$ for a video is the sum of activations within the effect mask over all $N$ frames:
\begin{equation}
     C_{b} = \frac{\sum_{f=1}^{N} \sum_{p}(\bm{S}_{f,b} \odot \bm{M}^e_f)}{
     \sum_{b=1}^{B}\sum_{f=1}^{N} \sum_{p}(\bm{S}_{f,b} \odot \bm{M}^e_f)
     }.
    \label{eq:activation}
\end{equation}
By averaging these scores over the full dataset, we obtain an architectural sensitivity profile that highlights which blocks are most responsive to associated effects (Fig.~\ref{fig:analysis_plot}, bottom).


Based on the block-wise contribution scores, which we plot in Fig.~\ref{fig:analysis_plot}, we partition the inpainting model into three distinct stages using the curve's troughs as clear delineators. The averaged $\bm{S}_b$ maps for each stage, also visualized in Fig.\ref{fig:analysis_plot} on real-world videos, elucidate their functional roles. The initial stage, characterized by broad receptive fields, primarily encodes contextual scene information. Blocks in the middle stage most strongly capture the spatial structures of foreground-associated effects. Conversely, in the final stage, effect-related features are actively suppressed. This observation validates our initial hypothesis on the inpainting mechanism.


\subsection{Dual Experts Sampling}
\label{sec:pipeline}
Based on the above findings, we involve a novel dual expert sampling strategy by \textit{Effect Expert} and \textit{Quality Expert} sequentially to obtain the final omnimatte as shown in Fig.~\ref{fig:pipeline}~(b). Below, we provide the details for each part.

\noindent\textbf{Effect Expert} 
is a model specialized in fine-tuning the original video inpainting model to capture the effects. 
Instead of applying LoRA to all DiT blocks, \emph{Effect Expert} has only LoRA fine-tuning blocks exclusively on the final stages of the inpainting DiT as shown in Fig.~\ref{fig:pipeline}. 
The motivation lies in the finding from Fig.~\ref{fig:analysis_plot}, which is to prevent the LoRA modules trained for alpha prediction from interfering with the inpainting model's inherent ability to perceive and remove associated effects.
Formally, 
denote the DiT blocks in the inpainint model as $\Theta=\{L_1, L_2, \dots, L_{B}\}$,
our Effect Expert model $\diff^E$ now has trainable weights:
\begin{equation}
    \Theta_{E} := \{\mathcal{B}(L_b) \ |\  b \in (B - K, B]\},
\end{equation}
where $\mathcal{B}$ is the branch-out operation adding LoRA to the last $K$ DiT block.
%
To train the \textit{Effect Expert}, we modify the self-attention mask so that the inpainting query tokens do not attend to the effect-matting tokens. This ensures that the inpainting branch remains unaffected, while its intermediate representations provide stable guidance for the effect-capturing branch (see Fig.~\ref{fig:pipeline}, top-middle).

\begin{figure}[tp]
    \centering
    \includegraphics[width=\linewidth]{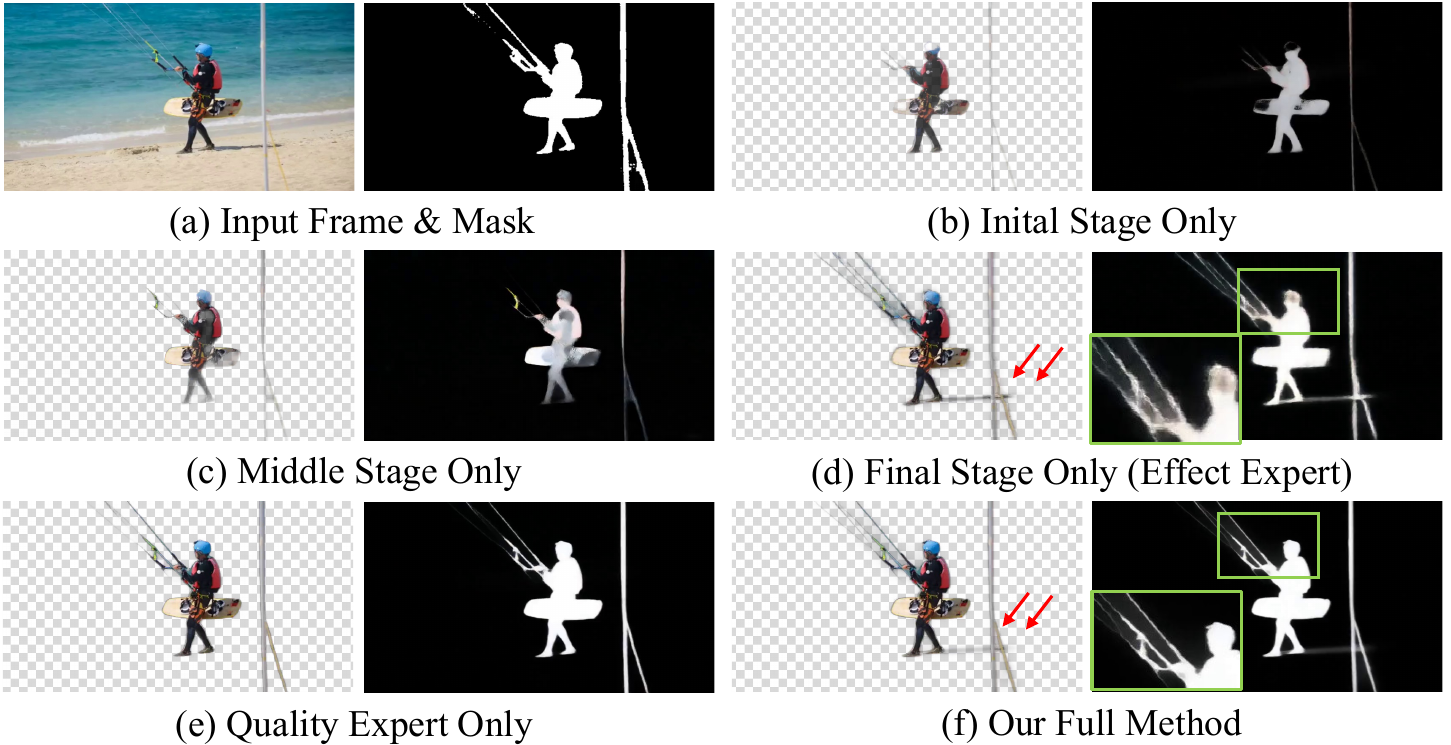}
    \vspace{-2em}
    \caption{\textit{\textbf{Dual Expert Sampling Strategy}}. We only add LoRA to the final stage to keep the prediction of the effects, and our full dual expert strategy further improves the matting quality.}
    \vspace{-1em}
    \label{fig:multi_ablate}
\end{figure}

\noindent\textbf{Quality Expert.}
As shown in Fig.~\ref{fig:multi_ablate}~(d), while the Effect Expert model significantly improves the preservation of effects, it comes at the cost of reduced precision in the overall matte shape compared to the single, full \emph{Quality Expert} with trainable weights $\Theta_O$:
\begin{equation}
    \Theta_{O} := \{\mathcal{B}(L_b) \ |\  b \in [1, B]\},
\end{equation}
To train \textit{Quality Expert}, we mask out the attention between the inpainting tokens and the matting tokens~(see Fig.~\ref{fig:pipeline}, bottom-middle). This design not only accelerates attention computation but also yields improved matting quality.

\noindent\textbf{Sampling.} 
We further introduce a \textit{Dual Expert Sampling Strategy} to achieve both high-fidelity effect prediction and accurate matte generation, and high computational efficiency. 
Our approach is inspired by the progressive nature of diffusion and flow-matching models, which generate content by gradually reducing noise over a series of timesteps, from high noise to low noise, to capture the target effects at early diffusion stages when the major content forms and refine details at later stages.
As illustrated in Fig.~\ref{fig:pipeline}~(b), we control the applied model of the diffusion timestep by the threshold $\tau$. Let $\diff^{E}$ and $\diff^{Q}$ denote the trained \textit{Effect Expert} model and \textit{Quality Expert} model, respectively,
the sampling progress at a timestep $t$ is determined as follows:
\begin{equation}
\diff =
    \begin{cases}
        \diff_{E} & \text{if } t > \tau, \\
        \diff_{Q} & \text{if } t \le \tau,
    \end{cases}
\label{eq:moe}
\end{equation}
where $\tau$ is set to 0.5 in our experiments suggested by Fig.~\ref{fig:tradeoff_curve}. These complementary experts operate in tandem across the high- and low-noise denoising stages, producing high-fidelity alpha mattes that accurately capture the associated effects, all without additional computational overhead.
\begin{figure*}[tp]
    \centering
    \includegraphics[width=1\textwidth]{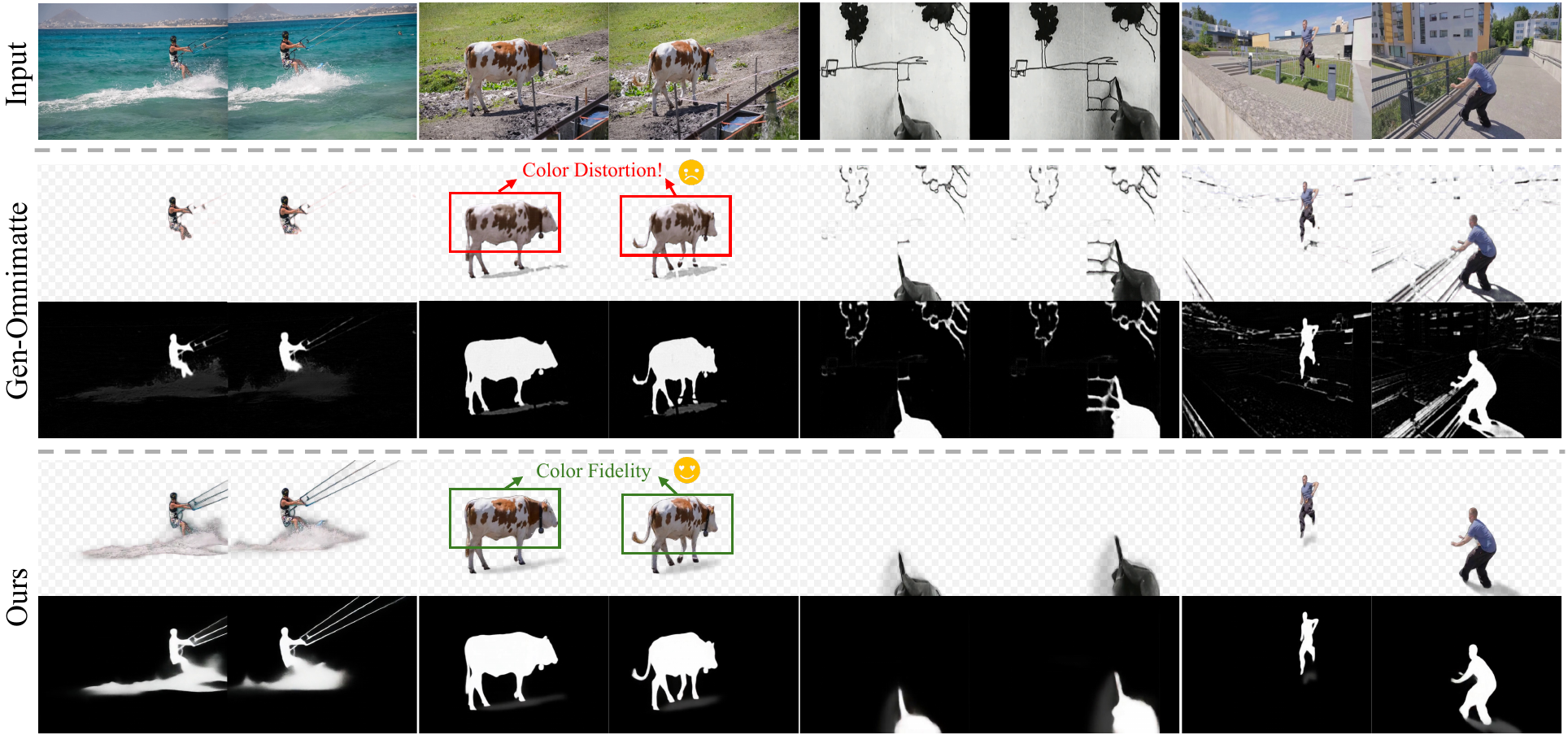}
    \vspace{-2em}
    \caption{Qualitative comparison with optimization-based omnimatte method~\cite{lee2025generative}. Please zoom in for a better view.}
    \vspace{-1.5em}
    \label{fig:omni_comp}
\end{figure*}

\section{Experiments}
\label{sec:experiment}

Our training data is synthetically generated by compositing foreground videos with high-quality alpha matte annotation~\cite{lin2021real} over background videos. The background videos are sourced from large-scale, high-resolution video datasets~\cite{lin2024open}, each accompanied by a descriptive caption. During composition, we apply a set of dynamic augmentations, including random resizing, rotation, and translation of the foreground, to enhance the diversity of the training samples. The foreground mask $M$ required by the inpainting model is derived directly from the ground-truth alpha matte $\alpha_{gt}$. To simulate foreground effects, we generate pseudo-shadows by applying a series of transformations (\eg, shearing, blurring, and color adjustments) to the foreground matte.

\begin{figure*}[tp]
    \centering
    \includegraphics[width=\textwidth]{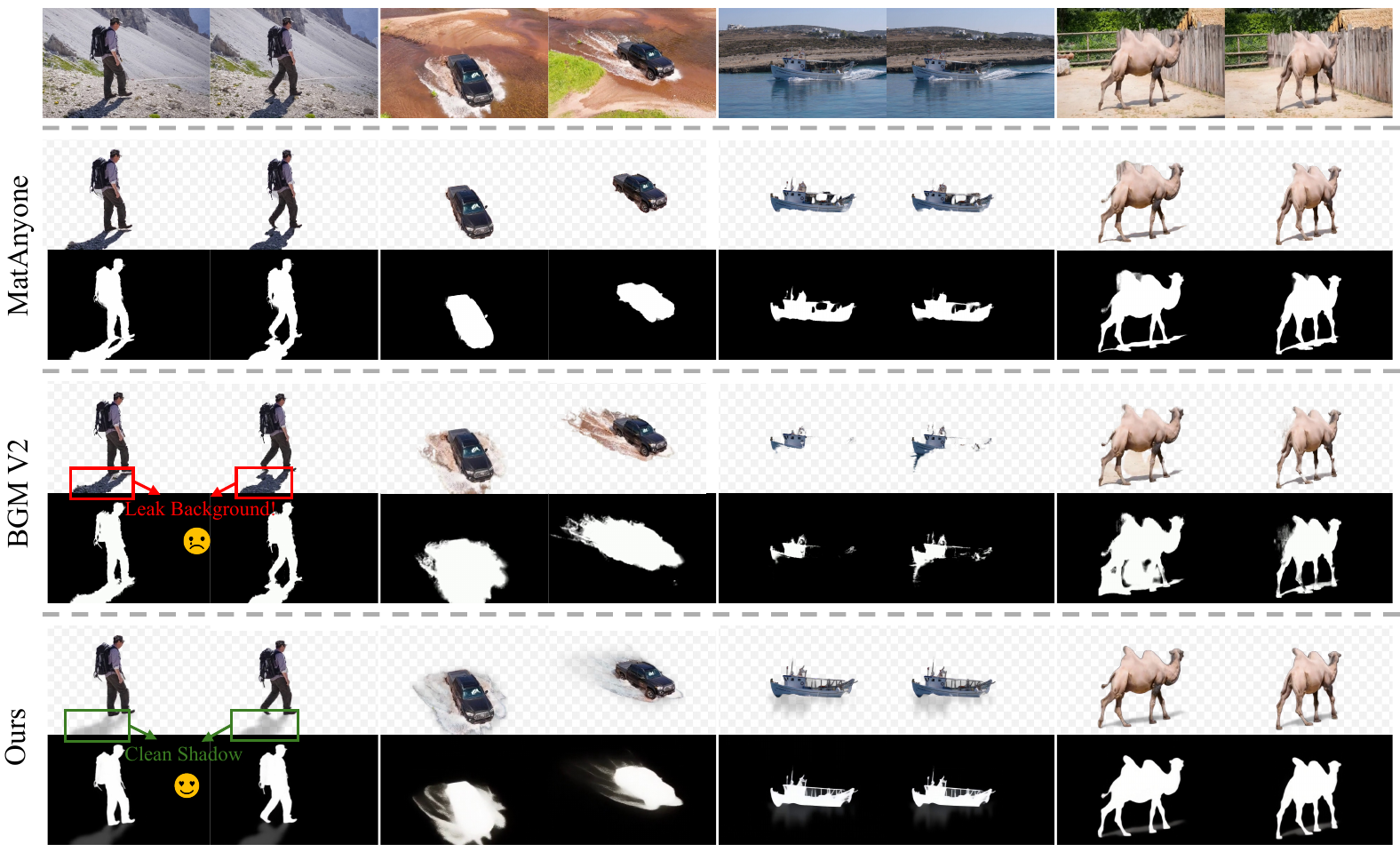}
    \vspace{-2em}
    \caption{Qualitative comparisons with learning-based matting method~\cite{lin2021real, yang2025matanyone}. Please zoom in for a better view.}
    \vspace{-1.5em}
    \label{fig:matting_comp}
\end{figure*}


For our experiments, we initialize our approach with a video inpainting model provided by Lee et al. \cite{lee2025generative}, which is built upon the open-source video generation framework from Wan et al. \cite{wan2025wan}. We then fine-tune this model to adapt it for the Omnimatte task. Our fine-tuning strategy incorporates a dual-expert architecture using LoRA, where the ranks for the Effect Expert and the Quality Expert are set to 128 and 64, respectively. For the dual-expert sampling strategy, we employ a default threshold of $\tau=0.5$. The entire model is trained end-to-end for 8,000 iterations using the AdamW optimizer \cite{loshchilov2017decoupled} with a learning rate of $1\times10^{-3}$. The training is conducted on two H100 GPUs.

\subsection{Comparison with State-of-the-Art Methods}
\label{sec:comparison}

\noindent\textbf{Optimization-based Omnimatte Prediction.} After generating the background with the same removal model we use, we employ the Omnimatte Optimization method from Gen-Omnimatte~\cite{lee2025generative} to decompose the foreground. Following the official recommendations, we set the number of optimization steps to 4000 and enabled detail transfer~\cite{lu2020layered}. Note that this is a test-time optimization method similar to the previous approaches~\cite {lu2021omnimatte, lin2023omnimatterf}, requiring several minutes of optimization for each foreground layer, which is over an order of magnitude slower than our approach. 

\noindent\textbf{Learning-based Matting Prediction.} We establish two categories of learning-based matting baselines. First, for background-matting methods, we use the output of our removal model to guide BGMv2~\cite{lin2021real}. Second, for mask-guided methods, we generate an initial mask for the foreground and its effects by integrating the segmentation model~\cite{ravi2024sam2} with SSIS-v2~\cite{Wang_2022_TPAMI}. This mask is then used to prompt the video matting model~\cite{yang2025matanyone} for prediction.
While contemporary learning-based methods are considerably more efficient than Gen-Omnimatte, they typically fail to preserve the object's associated effects, and even with auxiliary post-processing, the results are often suboptimal.

\noindent\textbf{Qualitative Comparison.} Fig.~\ref{fig:omni_comp} presents our comparison against the optimization-based method. Gen-Omnimatte~\cite{lee2025generative} suffers from color distortion, leading to poor foreground decomposition. Furthermore, it fails in scenarios with rapid background changes, resulting in significant background bleeding into the foreground layer. In contrast, our method overcomes these shortcomings while drastically reducing the inference time from several minutes to under 10 seconds. Fig.~\ref{fig:matting_comp} shows the comparison against dedicated matting methods. When capturing associated foreground effects, both matting baselines tend to predict excessively high alpha values in the effect regions. This results in the extracted foreground being contaminated with colors from the original background. Moreover, baselines exhibit poor generalization to non-human categories due to the missing of the generative priors, leading to significant artifacts and low-quality mattes.

\noindent\textbf{Quantitative Comparison and Human Evaluation.}
To objectively evaluate the performance of EasyOmnimatte in the absence of labeled test data, we design two quantitative experiments and incorporate a human evaluation study. The quantitative experiments include calculating a series of perceptual losses of the recomposed video against the original, as well as batch compositing onto numerous backgrounds and calculating the effect on the background video distribution. These experiments are based on the viewpoint that high-quality layered results can prevent the loss of original video information and allow for more harmonious synthesis into new backgrounds. More details can be found in the supplementary. We also conduct the human evaluation to assess the layering effect based on three criteria: foreground integrity, effect harmony, and temporal consistency. In our experiments, 28 users participated with 20 videos to study, resulting in 1,680 opinions.

\begin{table}[t]
    \centering
    \caption{\textit{\textbf{Quantitative comparion with baselines.}}  Notice that the BGM V2 and MatAnyone are not specifically for associated effects. $\dagger$ indicates cooperation with the shadow detection method~\cite{Wang_2022_TPAMI}.
    }
    \resizebox{\linewidth}{!}{
    \begin{tabular}{lcccc}
    \toprule
     \textbf{Method} & \textbf{PSNR} ($\uparrow$) & \textbf{SSIM} ($\uparrow$) & \textbf{WE}($\downarrow$) & \textbf{FVD} ($\downarrow$) \\
    \midrule
    \rowcolor{gray!20} BGM V2 \cite{lin2021real}    & \textbf{26.61} & \underline{78.78} & 101.04 & 168.31 \\
     \rowcolor{gray!20} MatAnyone$\dagger$ \cite{yang2025matanyone}  & 26.12 & 78.68 & \textbf{100.46} & 146.44 \\
    Gen-Omnimatte \cite{lee2025generative}  & 24.35 & 69.36 & 101.33 & \underline{116.32} \\
    EasyOmnimatte(Ours) & \underline{26.23} & \textbf{78.83} & \underline{100.94} & \textbf{105.48} \\
    \bottomrule
    \end{tabular}
    }
    \label{tab:metrics}
\end{table}


\begin{table}[t]
    \caption{ \textbf{\textit{User Studies}}. Human evaluation of different video generation methods across multiple aspects. Scores range from 0 to 5, with higher scores indicating better performance. Bold values represent the best performance within each metric. $\dagger$ indicates cooperation with the shadow detection method \cite{Wang_2022_TPAMI}.
    }   
    \resizebox{\linewidth}{!}{
    \centering
    \begin{tabular}{lcccc}
    \toprule
        \multirow{2}{*}{\textbf{Method}} & \textbf{Overall} & \textbf{Foreground} & \textbf{Effect} & \textbf{Temporal} \\
        & \textbf{Score} $\uparrow$ & \textbf{Integrity} $\uparrow$ & \textbf{Harmony} $\uparrow$ & \textbf{Consistency} $\uparrow$ \\
    \midrule
    BGM-V2              & 2.26 & 1.78 & 2.96 & 2.04 \\
    MatAnyone$^\dagger$          & 2.82 & 2.83 & 2.60 & 3.02 \\
    Gen-Omnimatte           & 2.85 & 2.45 & 3.36 & 2.74 \\
    EasyOmnimatte(Ours)          & \textbf{4.08} & \textbf{4.07} & \textbf{3.97} & \textbf{4.21} \\
    \bottomrule
    \end{tabular}
    }
    \vspace{-1em}
    \label{tab:user_study}
\end{table}


\subsection{Ablation Study}
\label{sec:ablation}

\begin{figure}[tp]
    \centering
    \includegraphics[width=\linewidth]{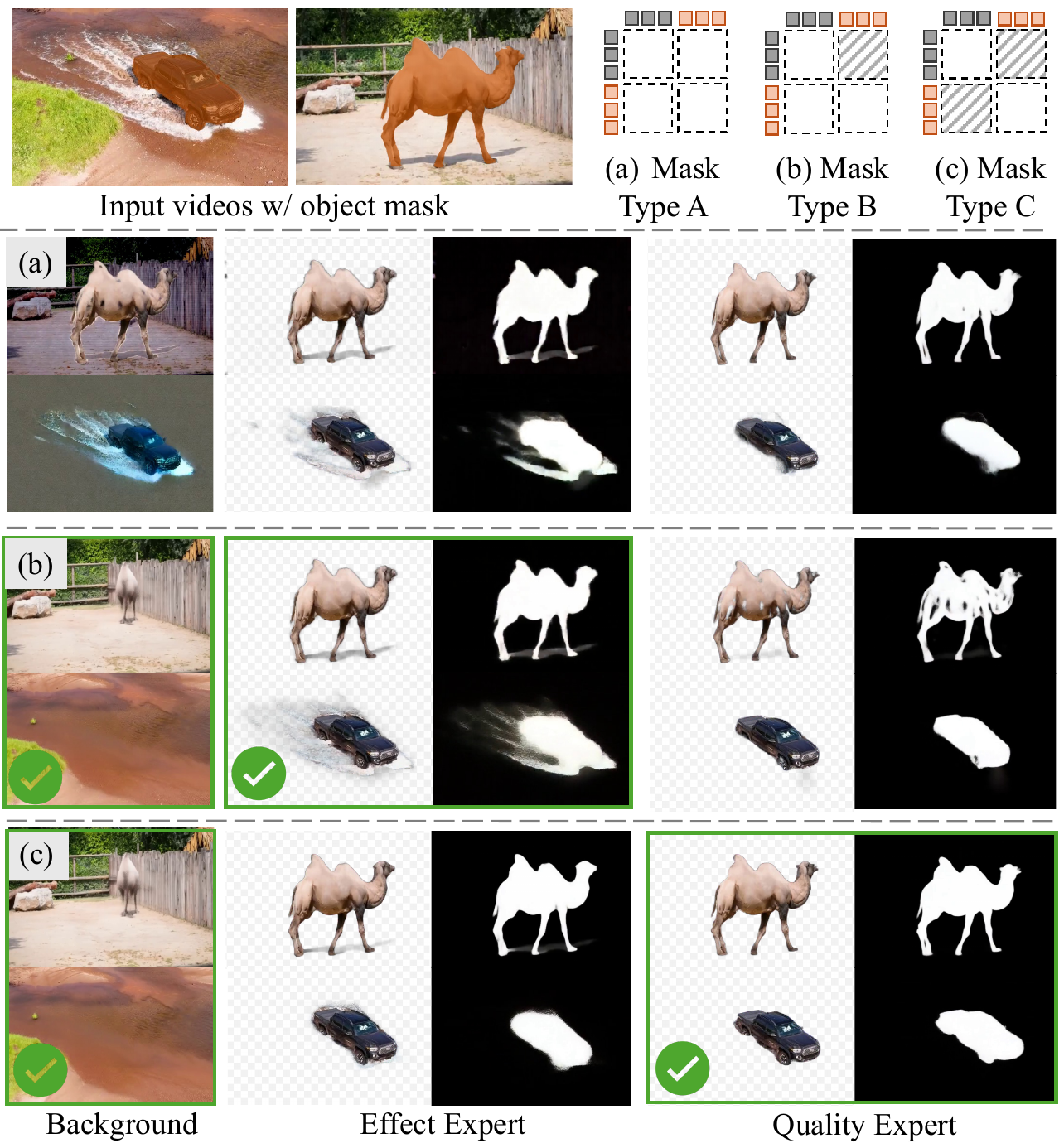}
    \vspace{-2em}
    \caption{\textbf{\textit{Ablation of Attention Masking}}. Training without the correct self-attention mask configuration renders both the Effect Expert and the Matting Expert ineffective. We use Mask type B and Mask type C for Effect Expert Quality Expert, respectively. }
    \vspace{-1.8em}
    \label{fig:mask_ablate}
\end{figure}

\noindent\textbf{Effectiveness of different attention masking strategies.} 
Figure~\ref{fig:mask_ablate} shows the results under different self-attention mask configurations. As shown in (a), allowing unrestricted communication between the branch tokens during inference leads to a corrupted background prediction, which in turn degrades the final foreground decomposition. Conversely, as seen in (b), preserving only the alpha-to-background attention path enhances the effect expert to perceive object-associated effects. However, this masking setup fails to resolve the poor quality of the alpha matte predicted by the Effect Expert. We attribute this issue to the masking strategy itself, rather than to the partial application of LoRA on only a subset of the Effect Expert's blocks. This is evidenced by the Quality Expert, which, despite being fine-tuned with LoRA across all blocks, also yields unsatisfactory results under the identical mask configuration. Therefore, our final strategy is to train the branches in complete isolation and then use the mask in (c) during inference. With this masking constraint, training the Quality Expert is equivalent to fine-tuning the inpainting model. Without being influenced by the frozen inpainting branch, the Quality Expert can quickly fit the matting data and form high-quality alpha mattes. However, the trade-off is that the Quality Expert remains incapable of perceiving effects, further highlighting the expert sample strategy.

\noindent\textbf{Ablation of Branch DiT Placement.} We experiment with modifying the Branch DiT Block to different inpainting stages. Our finding is consistent with the conclusion from our layer-wise analysis: applying LoRA to the later layers, which are responsible for suppressing effects, yields the most significant improvement in effect preservation.

\noindent\textbf{Validation of the Dual Expert Strategy.} 
We compare our full method against two primary variants: (d) using only the Effect Expert for the entire generation process, and (e) using only the Quality Expert. The results show that the Quality Expert achieves high precision on the main foreground but fails on effects, while the Effect Expert is strong on effects but sacrifices boundary precision. Our full model, which combines them via the hybrid diffusion sampling strategy, combines both of the advantages.

\begin{figure}[tp]
    \centering
    \includegraphics[width=0.9\linewidth]{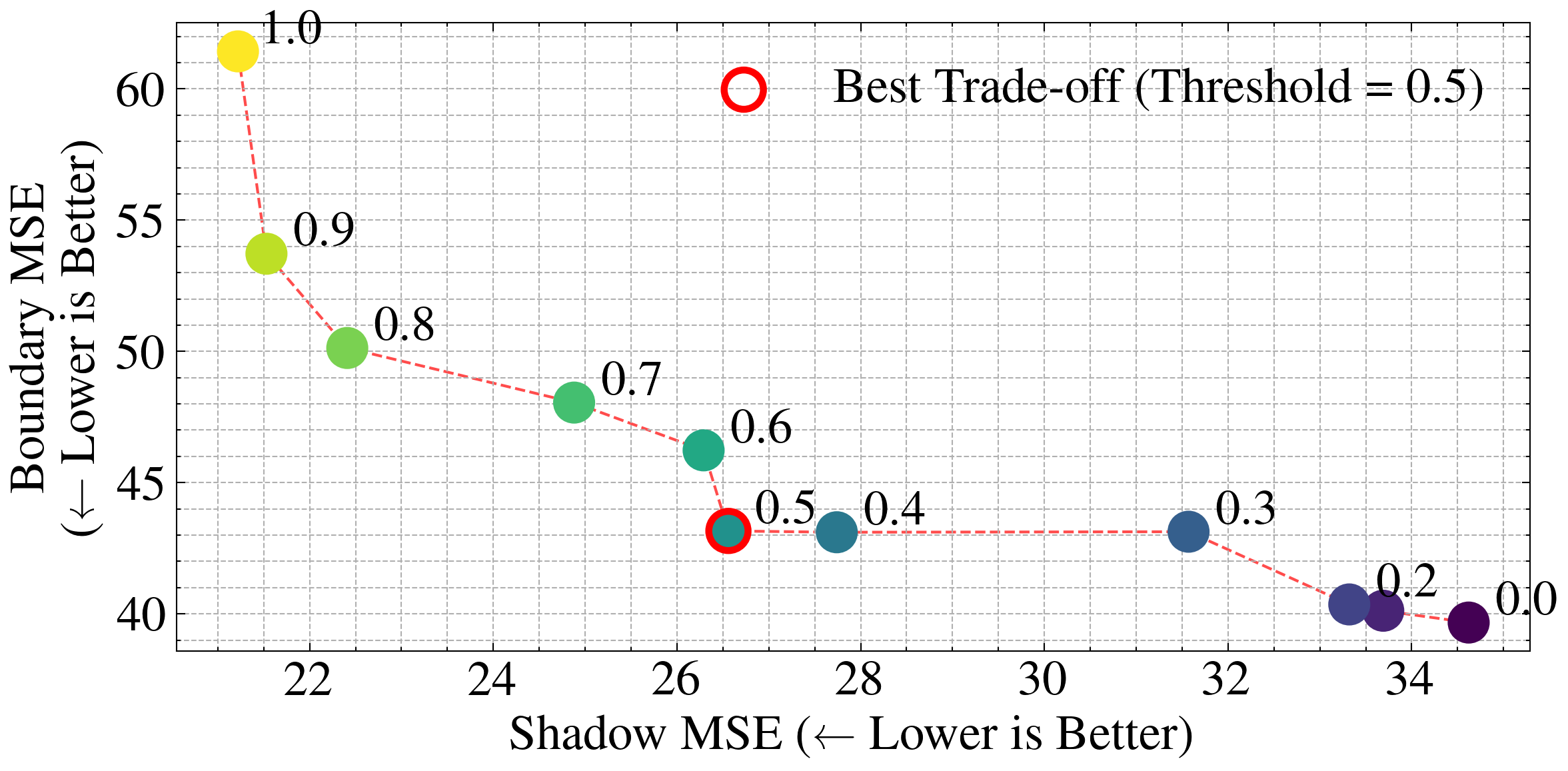}
    \vspace{-1em}
    \caption{Effect of varying \textit{\textbf{threshold $\tau$}}.By default, we set $\tau=0.5$ for a balanced outcome, but users can decrease it for higher matte precision or increase it for more prominent associated effects.}
    \label{fig:tradeoff_curve}
    \vspace{-1em}
\end{figure}

\noindent\textbf{Effect of different timestep threshold $\tau$.} To further investigate the impact of the $\tau$ in our Dual Experts Sampling strategy, we perform a quantitative analysis on a synthetic validation set. This set is generated by augmenting the test set of VideoMatte240K~\cite{lin2021real} with the pseudo-shadow augmentation method. We use the Mean Squared Error~(MSE) metric to separately assess the quality of the predicted foreground subject and the captured effects. The results of this analysis are visualized in Figure~\ref{fig:tradeoff_curve}.

\section{Conclusion}
\label{sec:conclusion}
We introduce EasyOmnimatte, a novel, end-to-end framework for video omnimatte, 
that is capable of capture complex foreground-associated effects with ease. 
Our key idea is to repurpose a powerful video inpainting model that has been proofed to be capable of perceiving the associated effects.
Nonetheless,
a trivial setup that directly apply LoRA to all DiT blocks fails to capture the effects.
We analyze the source of this failure and, consequently, propose the Dual Expert matting strategy. This strategy successfully captures the associated effects while simultaneously producing high-quality alpha matting. Our work establishes a new and highly effective paradigm for adapting large-scale generative models to complex, detailed video decomposition tasks.

\noindent\textbf{Limitations and Future Work.} 
While our method demonstrates superior video omnimatte performance, it is intrinsically linked to the capabilities of the base inpainting model. For future work, we will explore the application of our block-wise analysis and adaptation framework to more advanced and larger-scale generative models, pushing the boundaries of video omnimatte.

{\small
\bibliographystyle{ieee_fullname}

}

\clearpage \appendix
\label{sec:appendix}

This document provides supplementary details for our main paper. We begin with the preliminaries of our method (Sec.~\ref{sec:app_preliminaries}), followed by more details on the training data production (Sec.~\ref{sec:app_train_detail}), specific settings for our Block-wise Analysis (Sec.~\ref{sec:app_analysis_detail}), and the full experimental evaluation protocol (Sec.~\ref{sec:app_evaluation}). Furthermore, we present additional ablation studies (Sec.~\ref{sec:app_ablation}) and an analysis of failure cases (Sec.~\ref{sec:app_failure}). Full video comparisons are available on our project page.

\section{Prelinimaries}
\label{sec:app_preliminaries}

\noindent\textbf{WAN2.1 Model Structure.} Our work is built upon WAN2.1~\cite{wan2025wan}, a video diffusion model constructed as a stack of customized Diffusion Transformer (DiT) blocks~\cite{peebles2023scalable}+. Each block is composed of a visual self-attention layer for modeling spatio-temporal relationships within the video, and a cross-attention layer to incorporate textual conditioning, although text prompts are not used in our specific application.

To adapt this architecture for the video inpainting task, we follow a similar input formulation to Gen-Omnimatte~\cite{lee2025generative}. The process is as follows: first, the input video $\bm{V}$ is encoded into a latent representation $\mathbf{z}_v$ using a pre-trained Variational Autoencoder (VAE). This video latent is then concatenated along the channel dimension with the downsampled frame-wise binary mask $\bm{M}$ and a noise latent of the same spatial dimensions. The concatenated tensor is then passed through a linear projection layer to compress its channel dimension, forming the final sequence of visual tokens. These tokens are then duplicated and concatenated to serve as the input to the DiT model for the diffusion process.

\noindent\textbf{LoRA and Training Objectives.} To efficiently fine-tune the model for the Omnimatte task, we integrate Low-Rank Adaptation (LoRA)~\cite{hu2022lora} into the model's self-attention layers. For a pre-trained weight matrix $W_0 \in \mathbb{R}^{d \times k}$, the update is represented by a low-rank decomposition $W_0 + \Delta W = W_0 + BA$, where $B \in \mathbb{R}^{d \times r}$ and $A \in \mathbb{R}^{r \times k}$ are trainable matrices with a low rank $r \ll \min(d, k)$. The forward pass is modified as:
\begin{equation}
    h = W_0 x + BAx.
\end{equation}
In our framework, the Branch DiT applies this LoRA computation selectively to the copied tokens that are designated to learn the alpha matte, leaving the original tokens to be processed by the frozen, pre-trained weights.

Our model is trained with direct supervision on the alpha matte prediction. We employ the standard flow matching objective~\cite{lipman2022flow}, which trains the model to predict the noise $\boldsymbol{\epsilon}$ added to a clean latent $\mathbf{z}_0$ at timestep $t$. The loss function is defined as:
\begin{equation}
    \mathcal{L}_{\text{FM}} = \mathbb{E}_{t, \mathbf{z}_0, \mathbf{z}_1} \left[ || (\mathbf{z}_1 - \mathbf{z}_0) - \mathbf{v}_{\theta}(\mathbf{z}_t, t) ||^2 \right].
\end{equation}
where $\mathbf{z}_0$ is the VAE-encoded ground-truth alpha matte, $\mathbf{z}_1$ is the sampled Gaussian noise, $t$ is the diffusion timestep, and $\mathbf{v}_{\theta}$ is our network (with trainable LoRA parameters) that predicts the velocity from the noisy latent $\mathbf{z}_t$.

\section{Training Data}
\label{sec:app_train_detail}

While the sources and types of our input data are outlined in the main paper, this section provides a detailed description of our data augmentation process and presents visual examples of our training data.

\noindent\textbf{Data Augmentation.} We employ a temporally coherent video augmentation pipeline adapted from \cite{lin2021real, lin2022robust} to improve model robustness and generalization. This involves simulating camera motion by first embedding the sequence within a larger canvas using asymmetric padding (sampled from a uniform distribution $\mathcal{U}(0.3, 0.5)$ of the target width), followed by a smooth affine transformation. Rather than per-frame randomization, we interpolate between two affine states (A and B) using an easing function. To create pronounced lateral movement, the horizontal translations of A and B are set to be of opposite sign with magnitudes ranging from 15-30\% of the image width, while rotation ($\pm 5^{\circ}$), scale (0.95-1.05), and shear ($\pm 3^{\circ}$) are varied subtly.

A crucial component of our data synthesis is the addition of realistic associated effects. We focus on simulating shadows, as they are one of the most common and challenging effects to separate. For a given foreground alpha matte $\mathbf{\alpha}$, we generate a shadow matte $\mathbf{\alpha}_s$ by applying strong vertical compression (to 10-30\% of original height), significant horizontal shear ($30^{\circ}-60^{\circ}$), semi-transparent rendering (30-70\% opacity) and a final blur. This shadow matte is then used to darken the corresponding region on the background video before the final foreground is composited on top. This process forces the model to learn to identify and separate regions that are visually part of the foreground layer but are not captured by the original object mask.

\begin{figure}[tp]
    \centering
    \includegraphics[width=\linewidth]{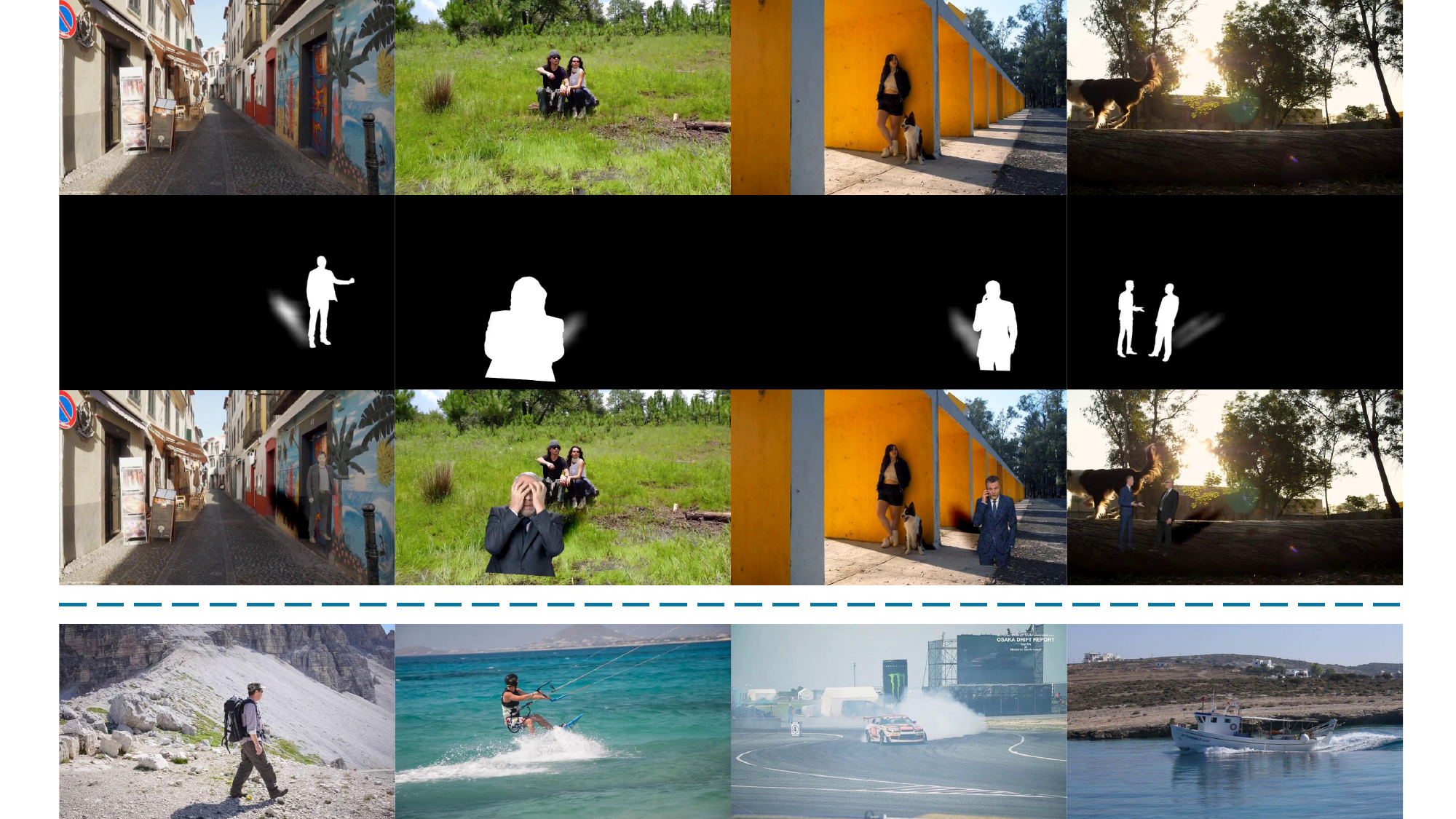}
    \caption{\textbf{Generalization from simple synthetic data to complex real-world scenes.} 
    Our method is trained exclusively on synthetic data (top) featuring only basic shadow effects. 
    Despite this, it successfully decomposes challenging in-the-wild videos (bottom) with a variety of unseen effects, including reflections and smoke. 
    This highlights the model's ability to bridge a substantial domain gap by learning a true separation principle.}
    \label{fig:data_examples}
\end{figure}

\noindent\textbf{Demonstration and Discussion.} In Fig.~\ref{fig:data_examples}, we provide a visual comparison between our synthetic training data and real-world videos from our test set. The top row showcases examples of our generated training frames, including the final composited video and the corresponding ground-truth alpha matte, which includes both the object and its simulated shadow. The bottom row presents frames from real videos, where the effects (e.g., natural shadows, reflections) are significantly more complex and subtle.

As is visually evident, a substantial \textit{domain gap} exists between our synthetic training data and the real-world test scenarios. Our training set, despite the augmentations, features clean-cut objects and simplified, programmatic shadows. In contrast, real videos contain complex lighting, soft and intricate effects, and various image artifacts. Despite this gap, our model demonstrates strong performance on these real-world examples, successfully isolating both the foreground object and its nuanced, naturally occurring effects. This robust generalization capability suggests that our end-to-end training approach has enabled the model to learn the fundamental, underlying logic of foreground-effect separation, rather than merely memorizing the specific characteristics of our synthetic dataset.

\section{Analysis Details}
\label{sec:app_analysis_detail}

\begin{figure}[tp]
    \centering
    \includegraphics[width=\linewidth]{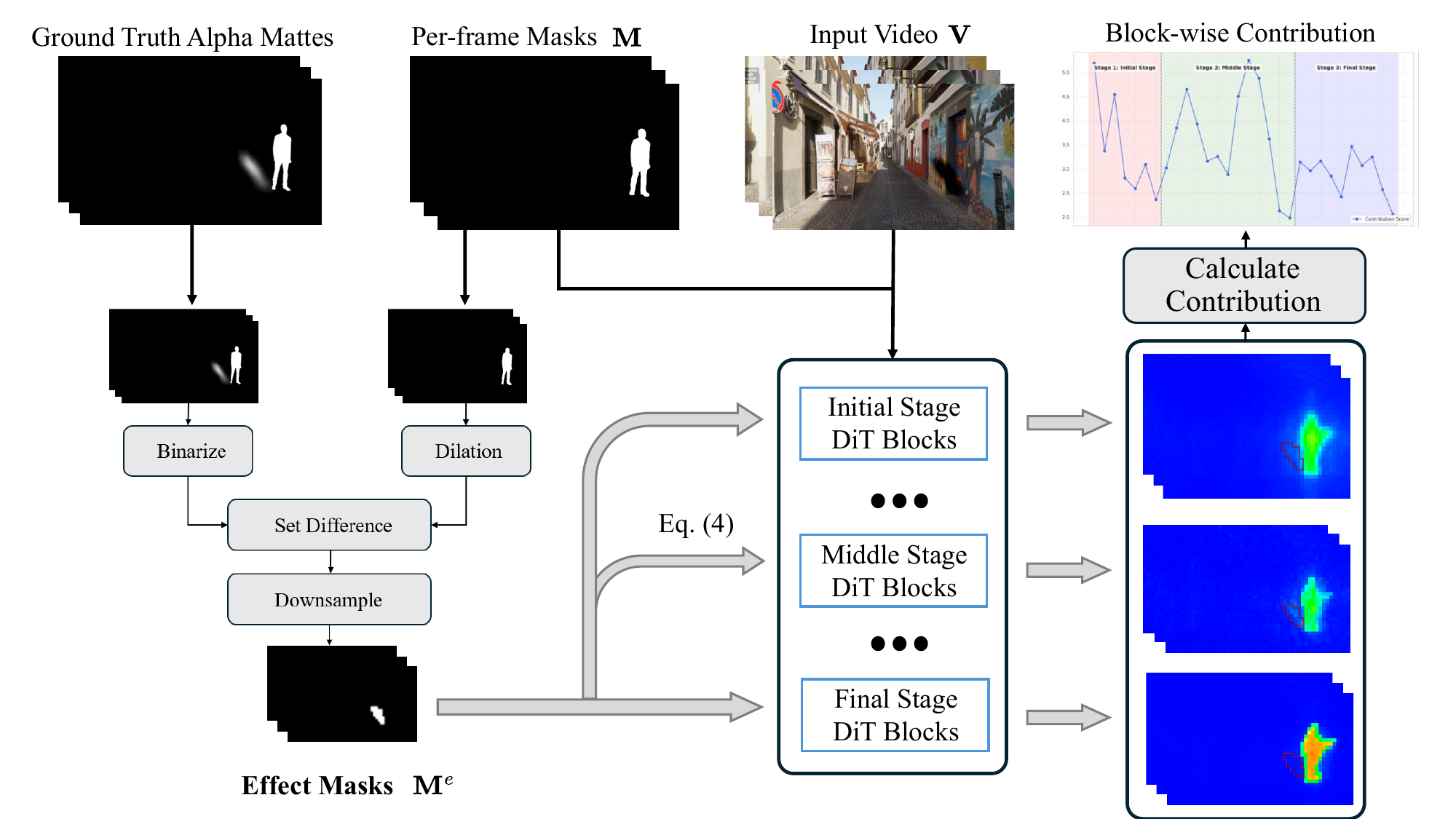}
    \caption{Pipeline of our Block-wise Contribution Analysis Method. The ground-truth alpha matte enables us to precisely localize the regions containing effects and subsequently compute the contribution score.}
    \label{fig:analysis_pipeline}
\end{figure}

Our block-wise analysis, presented in the main paper, was conducted on a test set of 1000 synthetic videos generated using the same pipeline as our training data but with held-out foreground and background clips. To obtain the effect mask $\bm{M}^e$ used in the analysis, we isolate the rendered shadow region from our synthetic data generation process. This provides a ground-truth spatial map of where the associated effect is located, allowing us to quantitatively measure each block's sensitivity to effect-related features. The whole pipeline is illustrated in Fig.~\ref{fig:analysis_pipeline}.

\section{Evaluation Protocols}
\label{sec:app_evaluation}

\noindent\textbf{Video Reconstruction Quality.} This protocol assesses the fidelity of the decomposition. We take a set of $P$ videos to carry out this experiment. The predicted foreground layer $\hat{\bm F}$ is composited back onto the predicted background layer using the predicted alpha matte $\hat{\mathbf{\alpha}}$. The quality is measured by comparing this reconstructed video with the original input video $\mathbf{V}$. A high-quality decomposition should allow for a near-perfect reconstruction.

We use three standard metrics: Peak Signal-to-Noise Ratio (PSNR), Structural Similarity Index Measure (SSIM), and Warping Loss to calculate temporal misalignments.

\noindent\textbf{Background Composition Plausibility.} This protocol evaluates how well the separated foreground layer can be composited onto novel, unseen backgrounds. We take a set of $P$ predicted foreground layers and compose them with a set of $Q$ different background videos. The key is that the separated foreground layer must be clean and free of artifacts from its original background to ensure a seamless new composition.

We use the Fréchet Video Distance (FVD) to quantitatively measure the quality of the newly generated videos. We compute the FVD between the set of our composited videos and the set of original background videos. A lower FVD score indicates that the distribution of the composited videos is closer to that of real videos, suggesting a higher-quality and more plausible decomposition.

\begin{figure*}[tp]
    \centering
    \includegraphics[width=\linewidth]{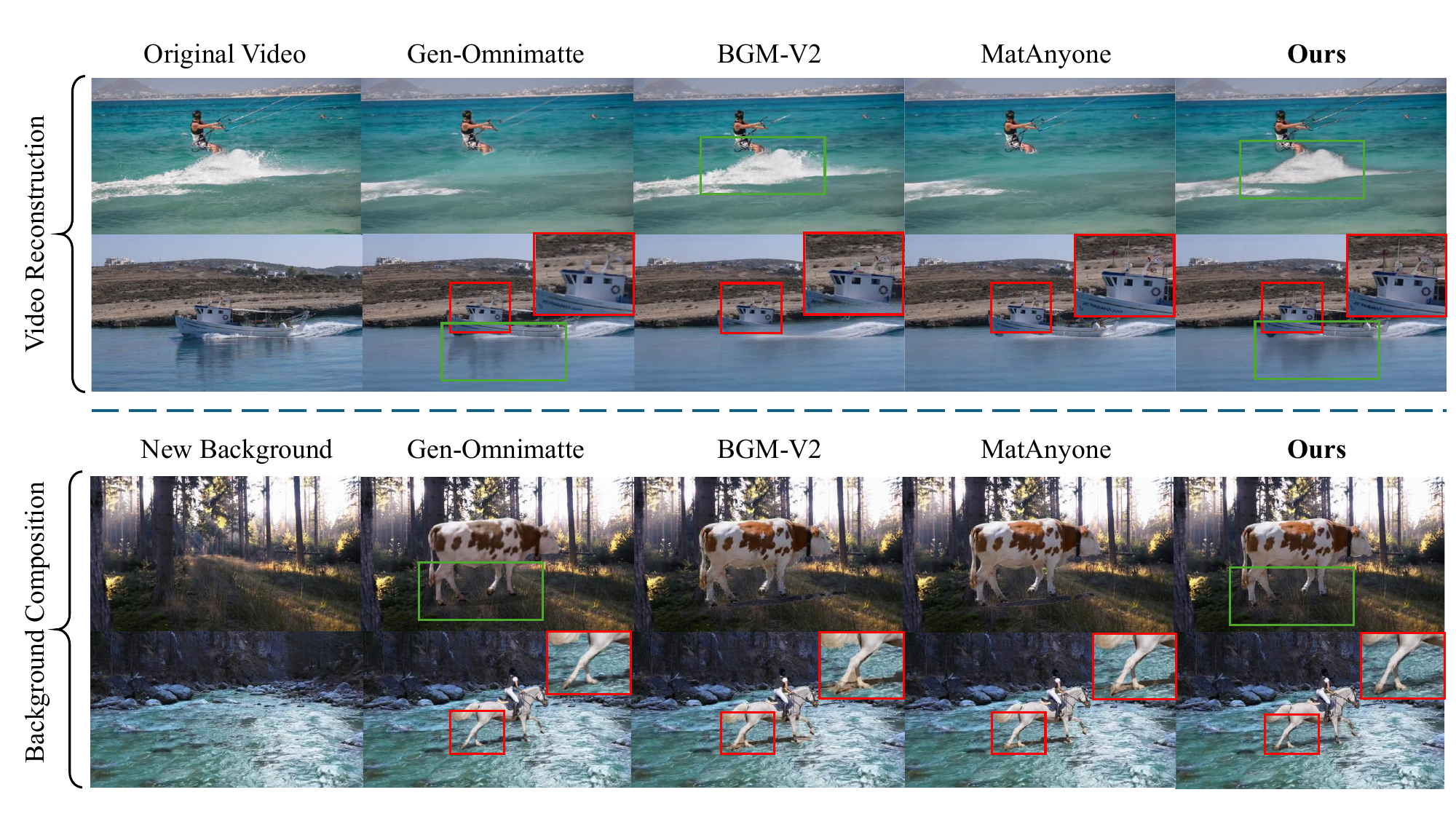}
    \caption{\textbf{Visualization of the Video Reconstruction and Background Composition.} The excellent performance of our method in both experiments stems directly from its superior ability to separate the foreground from its effects with high fidelity. We highlight correctly preserved effects in \textcolor{green}{green} boxes and magnify challenging details in \textcolor{red}{red} to demonstrate this capability. Zoom in for a better view.}
    \label{fig:quantitative_vis}
\end{figure*}

We present cases in Fig.~\ref{fig:quantitative_vis} to demonstrate the effectiveness of our method in the above experiments, where $P$ and $Q$ are 40 and 200, respectively. Our method demonstrates superiority in preserving the completeness and fine-grained details of the foreground. More critically, it possesses the capability to capture clean and intact associated effects, free from background artifacts.

\noindent\textbf{Human Evaluation.} To complement our quantitative metrics with a qualitative assessment of perceptual quality, we conducted a comprehensive user study. We recruited [Number of Participants, e.g., 30] participants with backgrounds in computer graphics and vision. The study was designed to compare our method against 4 leading baselines on a set of 20 challenging video sequences featuring a variety of objects and associated effects.

For each video sequence, participants were presented with a side-by-side comparison of the results from all methods, displayed in a randomized order to prevent bias. Participants were first briefed on the video decomposition task, and then they were shown how a clean separation is crucial for plausible compositing onto novel backgrounds.

Participants were asked to rate the quality of each method's output based on three predefined criteria, which were carefully explained to them beforehand:

\begin{enumerate}
    \item \textbf{Foreground Integrity:} This criterion assesses the completeness and color fidelity of the main object, emphasizing the absence of background color bleeding.
    \item \textbf{Effect Harmony:} This focuses on the integrity of the secondary effects, evaluating both their completeness and the plausibility of their rendered transparency.
    \item \textbf{Temporal Consistency:} This evaluates the temporal consistency and aesthetic quality of the final decomposition, penalizing artifacts such as temporal flickering, jagged boundaries, or other visual instabilities.
\end{enumerate}

For each criterion, participants provided a score on a 6-point scale, ranging from \textit{0 (very poor)} to \textit{5 (excellent)}. The final scores for each method were then averaged across all participants and video sequences to compute a \textbf{Overall Score} for each method aspect.

\begin{figure}[tp]
    \centering
    \includegraphics[width=\linewidth]{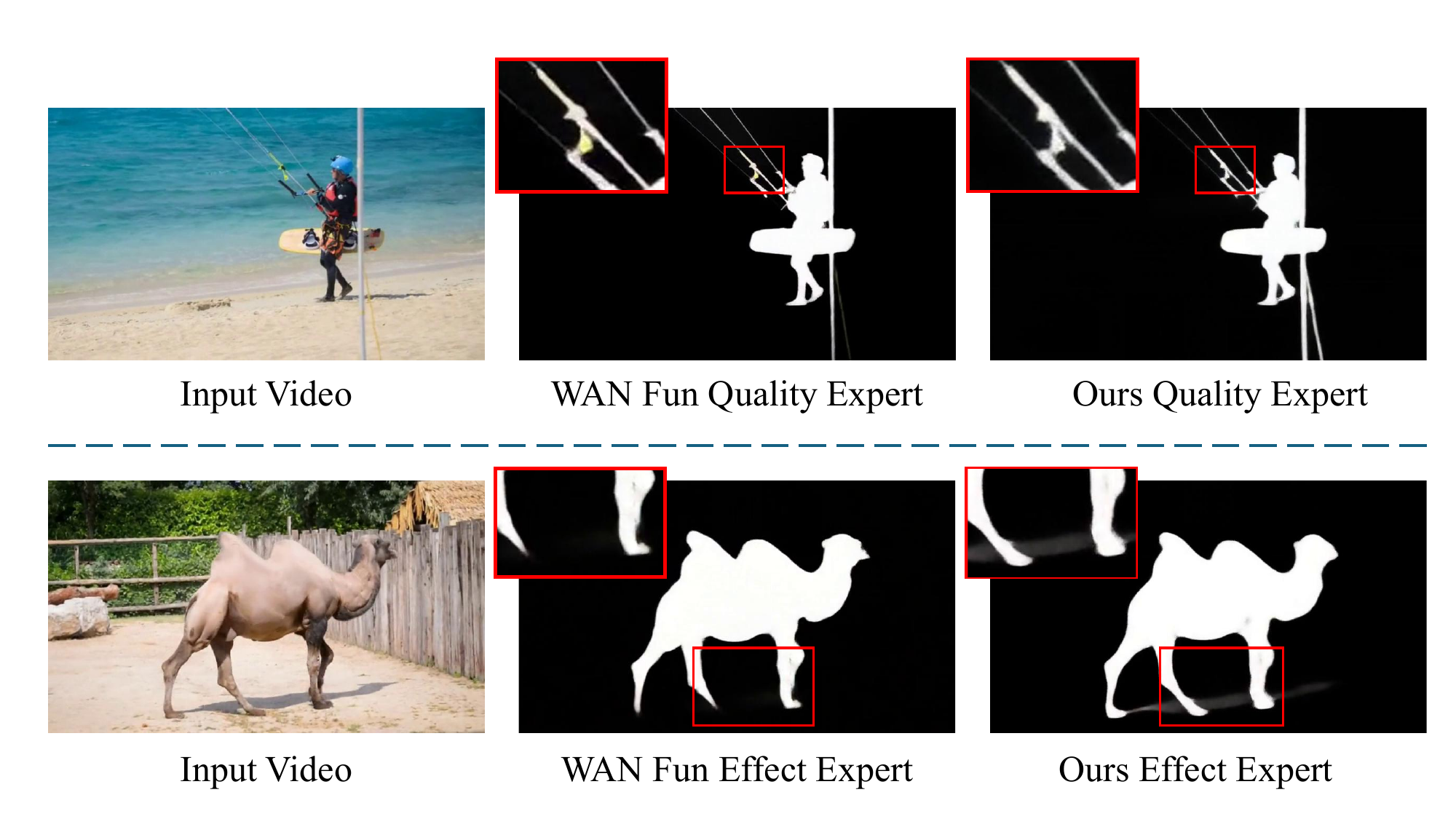}
    \caption{\textbf{Importance of Inpainting Pre-training.} Fine-tuning a general video model (WAN Fun) instead of an inpainting model leads to performance degradation. The Quality Expert (top) exhibits severe color bleeding artifacts in the alpha matte. The Effect Expert (bottom) fails entirely to capture shadows. This ablation confirms that an inpainting foundation is critical, especially for capturing associated effects.}
    \label{fig:more_ablate}
\end{figure}

\section{More Ablation Studies}
\label{sec:app_ablation}

In this section, we investigate the importance of the inpainting training of our base model. Specifically, we conduct an ablation study by fine-tuning the general-purpose conditional video generation model, WAN2.1 Fun, instead of our specialized video inpainting model. The primary goal is to determine whether the inherent knowledge of foreground removal is a prerequisite for successfully training a foreground decomposition model with our proposed strategy. The WAN2.1 Fun model shares the same input scheme with our inpainting base model but is trained for general conditional generation, not explicitly for object removal.

As shown in Fig.~\ref{fig:more_ablate}, we present the results of this ablation by training the WAN2.1 Fun model with our two expert modules:
\begin{itemize}
    \item \textbf{Training as a Quality Expert:} When fine-tuned to predict the alpha matte, we observe that the model can successfully learn to transfer to this new domain and generate a coarse alpha matte. However, the output suffers from significant artifacts, most notably a tendency to leak colors from the original video directly into the alpha matte, corrupting its purity. We hypothesize that this issue could potentially be alleviated with a much larger training dataset and extended training steps, but it highlights a fundamental difficulty for a general model to learn this task.

    \item \textbf{Training as an Effect Expert:} When trained to capture associated effects, the model completely fails to acquire this capability. The fine-tuned model does not learn to identify or isolate effects like shadows or reflections, indicating that this skill does not emerge naturally from a general video generation prior.
\end{itemize}

These experiments collectively demonstrate that under our current training strategy, fine-tuning a general generative model can roughly predict a primary alpha matte but is incapable of cleanly separating its associated effects. This deficit serves as strong evidence that the pre-training objective of \textit{object removal} is decisively important. 

\begin{figure}[tp]
    \centering
    \includegraphics[width=\linewidth]{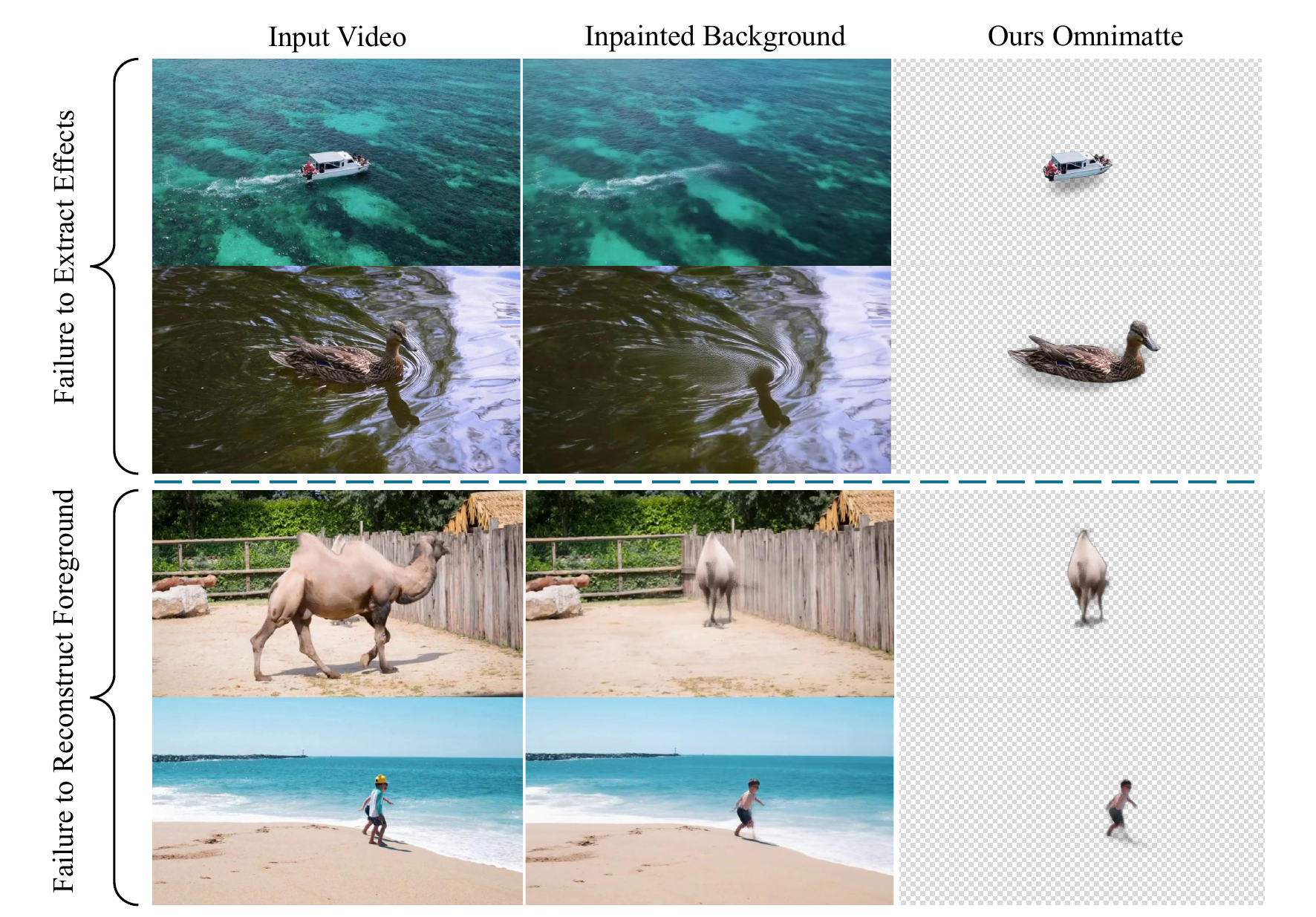}
    \caption{\textbf{Failure Cases.} Since our method is built upon an inpainting model, it naturally fails when the underlying inpainting model itself fails.}
    \label{fig:failures}
\end{figure}

\section{Failure Cases}
\label{sec:app_failure}

Our method's performance is inherently dependent on the capabilities of the underlying inpainting model. As illustrated in Fig.~\ref{fig:failures}, we observe two typical failure modes induced by this limitation:

\begin{itemize}
    \item \textbf{Failure to Extract Effects:} If the inpainting model fails to perceive an associated effect, it will treat that region as part of the true background. Consequently, our method cannot separate this effect. 
    \item \textbf{Failure to Reconstruct Foreground:} In cases where the inpainting model struggles to inpaint the object that is heavily occluded, its internal features for that region may become corrupted. This can lead to our method producing only a distorted foreground layer, as the decomposition is derived from these flawed features.
\end{itemize}

We believe that these limitations will be mitigated as the underlying video inpainting models become more powerful in their removal and completion capabilities. 

\end{document}